\documentclass[runningheads]{llncs}
\usepackage[T1]{fontenc}
\usepackage{newtxtext}       %
\usepackage{amssymb}
\usepackage[varvw]{newtxmath}       

\usepackage{makecell}
\usepackage{multirow}
\usepackage{float}
\usepackage{graphicx}
\usepackage{amsmath}
\begin{document}
\title{FocusOVD: Geometric and Contextual Feature Refinement for Open-Vocabulary Small-Object Detection}
\titlerunning{FocusOVD: Geometric and Contextual Feature Refinement}
%
%
\author{Hanlin Kou\inst{1}\orcidID{0009-0000-4502-3626} \and
Beiqi Bai\inst{1}\orcidID{0009-0000-5587-3843} \and
Li Chen\inst{1} \and
Youjia Jing\inst{2} \and
Shupan Li\inst{1} \and
Mingyuan Jiu\inst{1}}
\authorrunning{H. Kou et al.}
%
\institute{School of Computer Science and Artificial Intelligence, Zhengzhou University, Zhengzhou,China\\
\url{https://www7.zzu.edu.cn/csai/} \and
Sinosteel Zhengzhou Research Institute of Steel Wire Products Co., Ltd., China}
\maketitle              
\begin{abstract}
Open-vocabulary object detection aims to detect both seen and unseen categories by leveraging the semantic knowledge of vision-language models. However, existing methods often struggle with small objects due to insufficient preservation of high-resolution details, ineffective information aggregation across feature levels, and limited region-level discriminability. To address these challenges, we propose FocusOVD, an open-vocabulary object detector that enhances small-object representations without modifying the frozen VLM backbone. FocusOVD introduces three complementary components: a Reweighted Cross-scale Feature Fusion module, which learns normalized level-wise weights to adaptively balance shallow appearance details and deep semantic cues during feature pyramid aggregation; a Residual Pinwheel-shaped Gated Convolution module, which strengthens local geometric responses in high-resolution features; and a Residual Convolutional Gated Linear Unit module, which suppresses background-dominated activations and enhances RoI discriminability before prediction. Extensive experiments on COCO-OVD and LVIS-OVD demonstrate the effectiveness of FocusOVD. On COCO-OVD, FocusOVD achieves a novel-category AP of 36.3\%, outperforming CORA, UniDetector, and SAMP by 1.3\%, 1.1\%, and 1.5\%, respectively; on LVIS-OVD, it achieves an overall AP of 33.7\%, surpassing the strongest competing method by 0.6\%. These results demonstrate the strong generalization ability of FocusOVD for open-vocabulary small-object detection.

\end{abstract}
\section{Introduction}
Conventional closed-set detectors perform well on predefined categories but assume that the target vocabulary is known during training, limiting deployment in evolving environments \cite{ref1}. Open-vocabulary detection (OVD) addresses this limitation by leveraging pretrained vision-language models (VLMs) to recognize categories unseen during detector training \cite{ref3}. However, open-vocabulary small object detection (OVSD) remains challenging. Small objects occupy only a few pixels in the image, resulting in weak foreground information, and their detection demands finer-grained detail cues, more effective cross-scale interaction mechanisms, and more precise local representations \cite{ref2}. Moreover, general OVD methods are not specifically tailored to small object detection: high-resolution details tend to decay during feature propagation, and sparse RoI activation signals are easily overwhelmed by background responses \cite{ref1,ref2}, which renders small objects prone to missed detection or false positives.

Despite rapid progress in OVD, small-object recognition in open-vocabulary settings remains underexplored. To fill this research gap, we propose FocusOVD, a framework specifically designed for OVSD, targeting three bottlenecks: detail erosion along high-resolution pathways, insufficient level-wise reweighting between shallow spatial cues and deep semantic context, and weak region-level discrimination under sparse foreground evidence. Specifically, (1) Reweighted Cross-scale Feature Fusion (RCFF) is proposed to balance shallow appearance details and deep semantic context during feature pyramid aggregation by learning adaptive weights; (2) Residual Pinwheel-shaped Gated Convolution (RPGConv) is designed, which uses directional depthwise processing and a lightweight gate to enhance local geometric responses while preserving the original feature representation; (3) Residual Convolutional Gated Linear Unit (RCGLU) is introduced to refine RoI features through gated spatial modeling and a residual update, improving foreground-background discrimination before prediction.

The remainder of this paper is organized as follows: Section 2 reviews related work, Section 3 presents FocusOVD, Section 4 reports the experimental results, and Section 5 concludes the paper.

\section{Related Work}
\subsection{Open-Vocabulary Object Detection}
Open-vocabulary object detection (OVD) targets categories absent from detector training using large-scale vision-language models (VLMs). Existing approaches mainly fall into three categories: knowledge-distillation methods, region-text alignment methods, and frozen-VLM methods. Knowledge-distillation methods transfer VLM knowledge to a detector: ViLD distills CLIP image and text embeddings into region features, while HierKD \cite{ref21} transfers global and instance-level semantics through hierarchical distillation. These methods depend on auxiliary distillation objectives and teacher quality, adding training complexity. Region-text alignment methods learn region-language correspondence from grounding or caption data. RegionCLIP \cite{ref5} constructs region-text pairs with CLIP pseudo-labels for contrastive pretraining, whereas GLIP \cite{ref6} unifies detection and phrase grounding through joint pretraining. They require large-scale weakly labeled data and substantial pretraining. Frozen-VLM methods retain the pretrained backbone and train lightweight detection components. F-VLM \cite{ref7} directly exploits frozen CLIP features for localization and open-vocabulary classification, while CORA \cite{ref8} uses region prompting and anchor pre-matching to bridge image-level and region-level recognition. These methods preserve VLM generalization while reducing training cost. Recent work further improves region-language alignment and contextual modeling. DenseVLM \cite{ref27} decouples foreground and background semantics to reduce foreground bias in region-language alignment, while CODet \cite{ref28} incorporates visual and textual object co-occurrence knowledge to guide open-vocabulary detection.

\subsection{Small-Object Detection}
Small-object detection focuses on objects occupying few pixels, such as instances smaller than $32 \times 32$ pixels. Limited spatial resolution and feature degradation caused by downsampling make small objects difficult to detect. Existing methods address this problem primarily through enhanced multi-scale representation, contextual reasoning, or data and optimization strategies. FPN \cite{ref2}, PANet \cite{ref22}, and BiFPN \cite{ref15}, for instance, enhance multi-scale feature aggregation; SNIP \cite{ref24} and TridentNet \cite{ref25} adopt scale-aware training or feature processing, and PyramidBox \cite{ref26} exploits contextual information.

However, most small-object detectors assume closed-vocabulary recognition. In OVD, the frozen VLM backbone preserves semantic generalization but restricts adaptation: small-object features are easily diluted during pyramid aggregation and remain difficult to align with textual embeddings. Improving these representations without modifying the VLM backbone remains underexplored. FocusOVD addresses this gap without modifying the frozen VLM backbone by enhancing the detector-side FPN with cross-scale fusion and high-resolution local modeling, and refining RoI features before the Fast R-CNN head.
\section{Methodology}
\subsection{Overall Architecture}

\begin{figure}
\includegraphics[width=\textwidth]{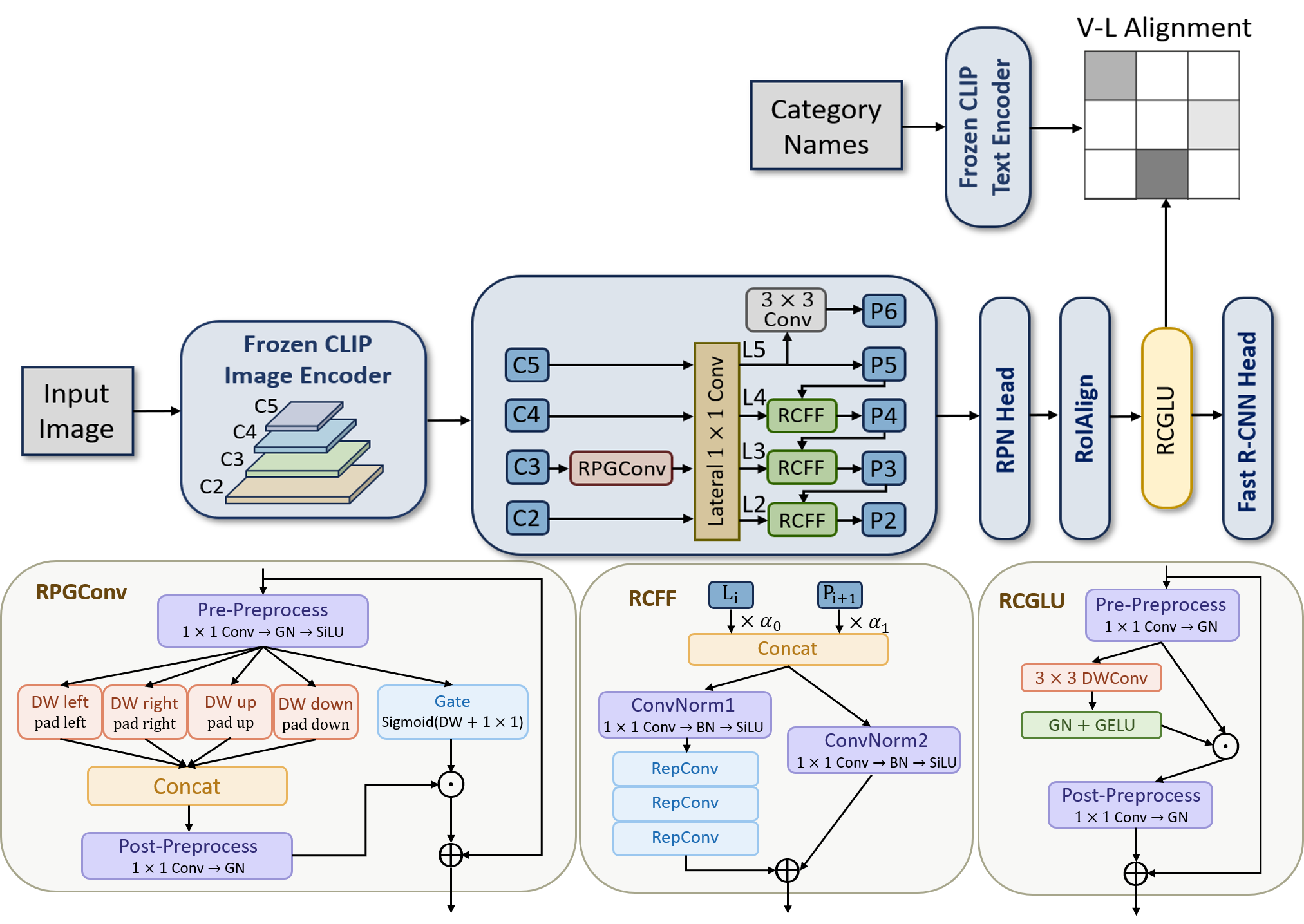}
\caption{ Overview of FocusOVD.} \label{focusovd}
\end{figure}

FocusOVD is a two-stage open-vocabulary detector \cite{ref1} built on a frozen CLIP-based vision-language model (VLM) \cite{ref3}, as shown in Fig.~\ref{focusovd}. Given an input image, the frozen image encoder extracts multi-level features $C_2$-$C_5$, while the text encoder maps category names into the shared vision-language embedding space. The text encoder is used to generate category embeddings for visual-text alignment. Then, these visual features are processed by a modified Feature Pyramid Network (FPN). In this network, RPGConv is inserted between high-resolution feature $C_3$ and $1 \times 1$ convolution to enhance local geometric details for small objects. RCFF replaces the conventional top-down fusion at the $P_2$, $P_3$ and $P_4$ levels, enabling adaptive aggregation of fine-grained and semantic information. The remaining layers follow the standard FPN design, producing multi-scale features $P_2$-$P_6$.

The RPN generates candidate boxes from the feature pyramid, and RoIAlign extracts fixed-size region features. RCGLU further refines these features before they are passed to the box head for classification and bounding-box regression. FocusOVD integrates frozen VLM semantics with enhanced high-resolution geometric modeling, cross-scale fusion, and RoI feature refinement to produce open-vocabulary detections.

\subsection{Residual Pinwheel-Shaped Gated Convolution}

To enhance local geometry while limiting changes to the frozen VLM representation, RPGConv is applied only to the high-resolution $C_3$ feature before its lateral projection, as shown in Fig.~\ref{focusovd}. Let $C$ denote the channel dimension of the input feature $C_3$. The compressed channel dimension is defined as
\begin{equation}
C_m = \max(\lfloor 0.5C \rfloor, 1).
\end{equation}
The input feature is first projected from $C$ to $C_m$ channels using a $1 \times 1$ convolution, followed by Group Normalization (GN) and the Sigmoid Linear Unit (SiLU) activation:
\begin{equation}
X = \mathrm{SiLU}\left(\mathrm{GN}\left(\mathrm{Conv}_{1 \times 1}(C_3)\right)\right) \in \mathbb{R}^{H \times W \times C_m}.
\end{equation}
GN normalizes feature channels within groups and does not depend on batch-level statistics, making it suitable for detection training with small batch sizes. The number of GN groups is determined by
\begin{equation}
N_{\mathrm{GN}}(C_m) = \max\left\{g \in \mathbb{N} \mid g \leq 32, C_m \bmod g = 0\right\}.
\end{equation}
The compressed feature $X$ is then processed by four asymmetric depthwise convolution (DWConv) branches arranged along the four cardinal directions. The left and right branches are horizontal branches using $1 \times 3$ kernels, while the up and down branches are vertical branches using $3 \times 1$ kernels. The four branches are defined as
\begin{align*}
X_l &= \mathrm{DWConv}_{1 \times 3}\left(\mathrm{Pad}_{left}(X)\right), \\
X_r &= \mathrm{DWConv}_{1 \times 3}\left(\mathrm{Pad}_{right}(X)\right), \\
X_u &= \mathrm{DWConv}_{3 \times 1}\left(\mathrm{Pad}_{up}(X)\right), \\
X_d &= \mathrm{DWConv}_{3 \times 1}\left(\mathrm{Pad}_{down}(X)\right).
\end{align*}
Where, $\mathrm{Pad}_{left}$, $\mathrm{Pad}_{right}$, $\mathrm{Pad}_{up}$, and $\mathrm{Pad}_{down}$ denote zero padding applied exclusively to the left, right, top, and bottom spatial boundaries, respectively.

The four directional responses are concatenated along the channel dimension:
\begin{equation}
X_{dir} = \mathrm{Concat}(X_l, X_r, X_u, X_d) \in \mathbb{R}^{H \times W \times 4C_m}.
\end{equation}
The horizontal branches capture directional responses along the left-right axis, whereas the vertical branches capture responses along the up-down axis. Their asymmetric receptive fields provide direction-aware local information for modeling object boundaries and corner structures.

To suppress background amplification, RPGConv uses a lightweight spatial-channel gate:
\begin{equation}
G = \sigma\left(\mathrm{DWConv}_{3 \times 3}(\mathrm{Conv}_{1 \times 1}(X)) + \mathrm{Conv}_{1 \times 1}(X)\right) \in (0, 1)^{H \times W \times C},
\end{equation}
where the convolutional projections produce $C$ channels. The enhanced response is
\begin{equation}
E = G \odot \mathrm{GN}\left(\mathrm{Conv}_{1 \times 1}(X_{dir})\right) \in \mathbb{R}^{H \times W \times C},
\end{equation}
where $\sigma(\cdot)$ is the sigmoid function and $\odot$ denotes element-wise multiplication. Finally, a learnable channel-wise scale provides a residual update:
\begin{equation}
C_3^{out} = C_3 + \beta \odot E, \quad \beta \in \mathbb{R}^C.
\end{equation}
The parameter $\beta$ is initialized to $10^{-3}$. During training, it is updated through back-propagation and the optimizer together with the other trainable parameters. The output preserves the spatial resolution and channel dimension of $C_3$ before entering the FPN lateral projection.

\subsection{Reweighted Cross-Scale Feature Fusion}

In a standard FPN, each top-down feature is obtained by adding a lateral feature to the upsampled feature from the adjacent level. Equal-weight summation can allow deep semantic responses to dominate shallow spatial details, weakening the boundary cues needed for small-object detection.To address this imbalance, we introduce Reweighted Cross-scale Feature Fusion (RCFF), which learns the relative contribution of the two inputs. Given the lateral feature $L_{i}$ and the upsampled feature $P_{i+1}$, the fusion weights are computed as
\begin{equation}
[\alpha_{0}, \alpha_{1}] = \mathrm{softmax}([w_{0}, w_{1}]),
\end{equation}
where $w_{0}$ and $w_{1}$ are learnable parameters initialized to 0.6 and 0.4, respectively. The softmax operation produces normalized weights that are positive and sum to one. The weighted features are concatenated along the channel dimension:
\begin{equation}
Z_{i} = \mathrm{Concat}\left(\alpha_{0} L_{i}, \alpha_{1} P_{i+1}\right).
\end{equation}
RCFF processes the concatenated feature $Z_{i}$ through a main branch and a lightweight residual branch. The main branch first projects $Z_{i}$ to the target channel width using a $1 \times 1$ convolution, Batch Normalization, and SiLU activation. Denoting this sequence as $\mathrm{CBS}_{1 \times 1}$, the projected feature is then processed by three sequential re-parameterizable convolution, denoted as $\mathrm{RepConv}_{3}$:
\begin{equation}
B_{i} = \mathrm{CBS}_{1 \times 1}(Z_{i}), \quad Z_{i}^{main} = \mathrm{RepConv}_{3}(B_{i}).
\end{equation}

\begin{figure}
\centering
\includegraphics[width=0.5\textwidth]{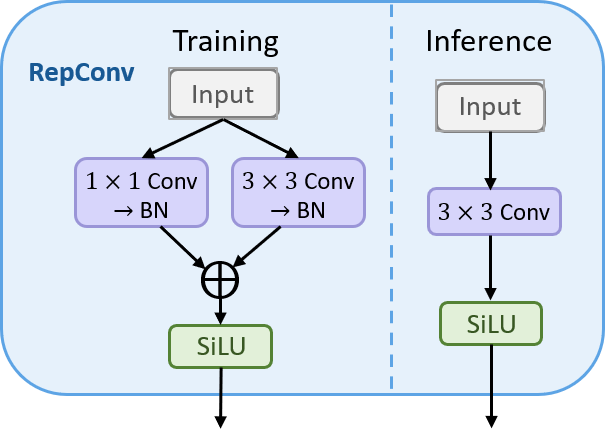}
\caption{Structure of the re-parameterizable convolution used in RCFF.}
\label{fig:repconv}
\end{figure}

The re-parameterizable convolution used in RCFF follows the multi-branch training and single-branch inference. The structure of the block is illustrated in Fig.~\ref{fig:repconv}. During training, each RepConv contains parallel $3 \times 3$ and $1 \times 1$ convolutional branches, each followed by Batch Normalization. If $\mathrm{CB}_{k \times k}$ denotes a $k \times k$ convolution followed by Batch Normalization, one re-parameterizable convolution is written as
\begin{equation}
\mathrm{RepConv}(B_{i}) = \mathrm{SiLU}\left(\mathrm{CB}_{3 \times 3}(B_{i}) + \mathrm{CB}_{1 \times 1}(B_{i})\right).
\end{equation}
During inference, the parallel convolution-Batch Normalization branches can be re-parameterized into an equivalent $3 \times 3$ convolution followed by SiLU activation.

The residual branch applies an independent $1 \times 1$ convolution, Batch Normalization, and SiLU activation:
\begin{equation}
Z_{i}^{res} = \mathrm{CBS}_{1 \times 1}(Z_{i}).
\end{equation}
The final RCFF output is obtained by summing the two branches:
\begin{equation}
P_{i} = Z_{i}^{main} + Z_{i}^{res}.
\end{equation}
To control computational overhead, RCFF is applied at the $P_{2}$, $P_{3}$ and $P_{4}$ levels, where the imbalance between semantic context and fine-grained spatial details has the greatest impact on small-object localization. The remaining FPN levels retain the standard fusion operation.

\subsection{Residual Convolutional Gated Linear Unit}

After RoIAlign, each proposal is represented by a fixed-resolution feature map. For small objects, this representation contains limited spatial evidence and may lose the local structure needed to distinguish foreground from clutter. We therefore insert a Residual Convolutional Gated Linear Unit (RCGLU) between RoIAlign and the Fast R-CNN head.

Let $F \in \mathbb{R}^{N \times S \times S \times C}$ denote the pooled RoI features. RCGLU first expands the channel dimension to $2C_h$, where
\begin{equation}
C_h = \max(\lfloor 2C \rfloor, 1).
\end{equation}
The expanded feature is obtained using a $1 \times 1$ convolution followed by Group Normalization:
\begin{equation}
F_z = \mathrm{GN}\left(\mathrm{Conv}_{1 \times 1}(F)\right) \in \mathbb{R}^{N \times S \times S \times 2C_h}.
\end{equation}
The expanded feature is split evenly along the channel dimension into a gate branch and a value branch:
\begin{equation}
(F_g, F_v) = \mathrm{Split}_{channel}(F_z), \quad F_g, F_v \in \mathbb{R}^{N \times S \times S \times C_h}.
\end{equation}
To make the gate spatially aware, the gate branch applies a depthwise $3 \times 3$ convolution, Group Normalization, and GELU activation:
\begin{equation}
\widetilde{F}_g = \mathrm{GELU}\left(\mathrm{GN}\left(\mathrm{DWConv}_{3 \times 3}(F_g)\right)\right).
\end{equation}
Same padding is used to preserve the spatial resolution of the RoI feature map. The gated feature is then obtained by element-wise modulation:
\begin{equation}
F_u = \widetilde{F}_g \odot F_v.
\end{equation}
The $1 \times 1$ convolution and Group Normalization project the modulated feature back to $C$ channels:
\begin{equation}
Y = \mathrm{GN}\left(\mathrm{Conv}_{1 \times 1}(F_u)\right) \in \mathbb{R}^{N \times S \times S \times C}.
\end{equation}
Finally, RCGLU applies a channel-wise residual update:
\begin{equation}
F_{out} = F + \gamma \odot Y, \quad \gamma \in \mathbb{R}^{1 \times 1 \times 1 \times C},
\end{equation}
where $\gamma$ is a learnable channel-wise scaling vector initialized to $10^{-2}$ and broadcast over all proposals and spatial locations. The output $F_{out}$ retains the spatial and channel dimensions of $F$ and is forwarded to the Fast R-CNN head.

\subsection{Training and Inference}

During training, the CLIP backbone remains frozen, while the feature pyramid, RPN, and detection head are jointly optimized with the standard detection objective:
\begin{equation}
L = L_{\mathrm{rpn}}^{\mathrm{cls}} + L_{\mathrm{rpn}}^{\mathrm{bbox}} + L_{\mathrm{roi}}^{\mathrm{cls}} + L_{\mathrm{roi}}^{\mathrm{bbox}},
\end{equation}
where $L_{\mathrm{rpn}}^{\mathrm{cls}}$ and $L_{\mathrm{rpn}}^{\mathrm{bbox}}$ measure proposal-stage objectness classification and bounding-box regression, respectively, while $L_{\mathrm{roi}}^{\mathrm{cls}}$ and $L_{\mathrm{roi}}^{\mathrm{bbox}}$ measure RoI-level category classification and bounding-box regression. The RPN classification loss $L_{\mathrm{rpn}}^{\mathrm{cls}}$ uses binary cross-entropy over positive and negative anchors from all pyramid levels. The RPN regression loss $L_{\mathrm{rpn}}^{\mathrm{bbox}}$ applies the Huber loss to the coordinate differences between predicted and target boxes only for positive anchors, with transition threshold $\delta = 0.111$. For the detection head, the classification loss $L_{\mathrm{roi}}^{\mathrm{cls}}$ uses softmax cross-entropy over sampled RoIs, with the background class assigned a weight of 0.9 and foreground classes a weight of 1.0. The box regression loss $L_{\mathrm{roi}}^{\mathrm{bbox}}$ is class-agnostic and applies the Huber loss only to foreground RoIs; background RoIs are excluded from the regression loss. We apply weight decay of $10^{-3}$ to trainable parameters, excluding normalization layers and the learnable temperature parameter.

The detection head does not learn a fixed closed-set classifier. Instead, each region feature is L2-normalized and compared with category text embeddings produced by the CLIP text encoder. The cosine similarities are divided by a learnable temperature to produce pre-softmax class scores, allowing the vocabulary to change by substituting the corresponding text embeddings at inference time.

During inference, we additionally use the frozen CLIP branch as an auxiliary classifier. RoI features are projected into the shared vision-language space through the fixed attention-pooling layer and matched against offline-generated category embeddings. The resulting similarities use a fixed temperature of 0.01. Let $p_{\mathrm{det},c}$ and $p_{\mathrm{clip},c}$ denote the detector and CLIP probabilities for category $c$, respectively. For each foreground category, we compute an unnormalized fused score as
\begin{equation}
\widetilde{p_c} = p_{\mathrm{det},c}^{1-\lambda} \cdot p_{\mathrm{clip},c}^{\lambda},
\end{equation}
where $\lambda = 0.35$ for base classes and $\lambda = 0.65$ for novel classes. Thus, the detector dominates base-category predictions, while the frozen CLIP branch contributes more strongly to novel categories. The background score is taken directly from the detector branch. After renormalization, score thresholding and NMS produce the final detections.

\section{Experiments}
\subsection{Experimental Setup}

Images are resized to 512 pixels with scale jittering in the range \([0.1, 2.0]\). The feature pyramid uses levels \(P_2\)-\(P_6\), and the RoI output size is \(7 \times 7\). The anchor size is 8, and at most 100 detections are retained per image. Category text embeddings are generated offline from category names and kept fixed throughout training and inference \cite{ref12,ref13,ref14}. Models are trained in float32 with batch size 4, while evaluation uses batch size 1. We use SGD with momentum and weight decay \(10^{-3}\), excluding normalization layers and the learnable temperature parameter. The learning rate is linearly warmed up from 0.0032 to 0.02 over the first 100 steps, followed by stepwise decay with milestones at 80\%, 90\%, and 95\% of training and a decay factor of 0.1. The background and foreground classification weights are 0.9 and 1.0, respectively. The fixed CLIP similarity temperature is 0.01, and the number of pre-NMS detections is fixed at 1,000.

\subsection{Evaluation on Open-Vocabulary Benchmarks}

We first evaluate FocusOVD on the COCO \cite{ref13} and LVIS \cite{ref12} open-vocabulary benchmarks. For COCO \cite{ref13}, we follow the standard split with 48 base and 17 novel categories. The model is trained using only base-category annotations and evaluated on all 65 categories. We report AP at an IoU threshold of 0.50 for novel categories (AP\(_{\text{novel}}\)), base categories (AP\(_{\text{base}}\)), and all categories (AP\(_{\text{all}}\)). To further evaluate scale-specific performance, we report AP\(_{\text{S}}^{\text{novel}}\), AP\(_{\text{M}}^{\text{novel}}\), and AP\(_{\text{L}}^{\text{novel}}\) for novel categories, corresponding to object areas smaller than \(32^2\), between \(32^2\) and \(96^2\), and larger than \(96^2\), respectively. Table~\ref{tab1} summarizes the open-vocabulary detection results on COCO-OVD.

\begin{table}[H]
\caption{Open-vocabulary detection results on COCO-OVD, including size-specific AP for novel categories.}
\label{tab1}
\centering
\small
\begin{tabular}{lcccccc}
\hline
Model & AP\(_{\text{novel}}\) & AP\(_{\text{base}}\) & AP\(_{\text{all}}\) & AP\(_{\text{S}}^{\text{novel}}\) & AP\(_{\text{M}}^{\text{novel}}\) & AP\(_{\text{L}}^{\text{novel}}\) \\
\hline
ViLD \cite{ref4} & 27.6 & 58.7 & 51.3 & 14.5 & 29.1 & 39.3 \\
RegionCLIP \cite{ref5} & 31.4 & 57.1 & 50.4 & 18.2 & 33.5 & 44.4 \\
OV-DETR \cite{ref17} & 29.4 & \underline{61.0} & 52.7 & 17.8 & 32.6 & 41.0 \\
Detic \cite{ref18} & 27.8 & 47.1 & 45.0 & 15.7 & 30.0 & 39.9 \\
CoDet \cite{ref10} & 30.2 & 52.4 & 46.7 & 17.5 & 32.8 & 41.1 \\
BARON \cite{ref20} & 34.0 & 60.4 & 53.5 & 20.3 & 36.5 & 46.3 \\
CORA \cite{ref8} & 35.0 & 44.2 & 43.5 & 21.9 & 38.0 & 47.7 \\
UniDetector \cite{ref9} & \underline{35.2} & 56.8 & 51.2 & \underline{22.4} & \underline{38.3} & \textbf{48.5} \\
SAMP \cite{ref23} & 34.8 & \textbf{61.4} & \textbf{54.2} & 21.9 & 38.2 & 47.1 \\
FocusOVD & \textbf{36.3} & 60.9 & \underline{54.5} & \textbf{24.3} & \textbf{38.9} & \underline{48.3} \\
\hline
\end{tabular}
\end{table}

\begin{figure}
\centering
\includegraphics[width=0.8\textwidth]{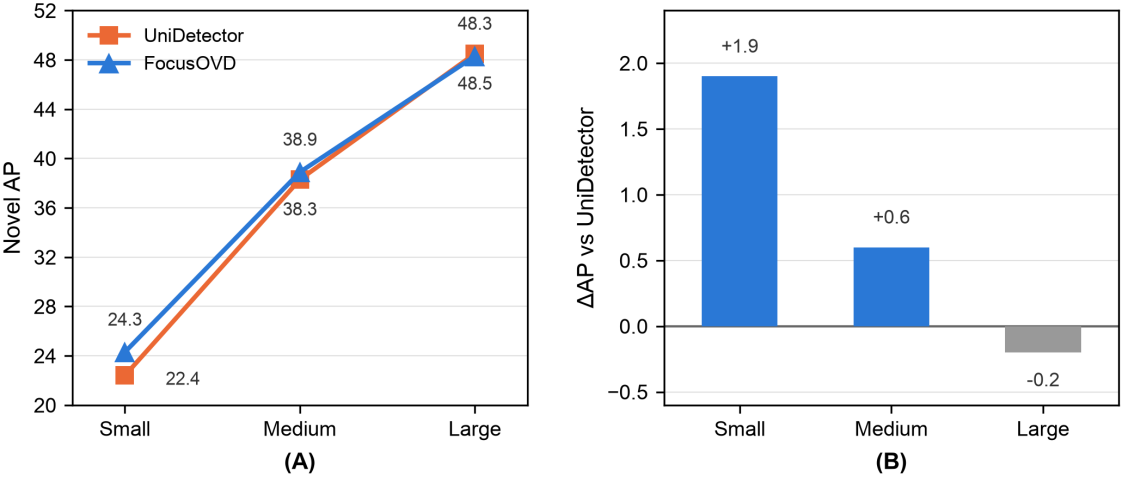}
\caption{Novel-category AP on COCO-OVD for UniDetector and FocusOVD across different object scales.}
\label{Novel_category}
\end{figure}

As shown in Table~\ref{tab1}, FocusOVD achieves 36.3\% AP\(_{\text{novel}}\), outperforming CORA, UniDetector, and SAMP by 1.3\%, 1.1\%, and 1.5\%, respectively. It also reaches 54.0\% AP\(_{\text{all}}\), ranking second among the compared methods and remaining competitive on base categories. For novel objects across different scales, FocusOVD obtains 24.3\% AP\(_{\text{S}}^{\text{novel}}\), 38.9\% AP\(_{\text{M}}^{\text{novel}}\), and 48.3\% AP\(_{\text{L}}^{\text{novel}}\). Compared with UniDetector, FocusOVD improves AP\(_{\text{S}}^{\text{novel}}\) and AP\(_{\text{M}}^{\text{novel}}\) by 1.9\% and 0.6\%, respectively, while obtaining a comparable result on large objects. Fig.~\ref{Novel_category} shows that the largest improvement over UniDetector occurs on small objects, where FocusOVD achieves a gain of 1.9\%.

\begin{figure}
\centering
\includegraphics[width=\textwidth]{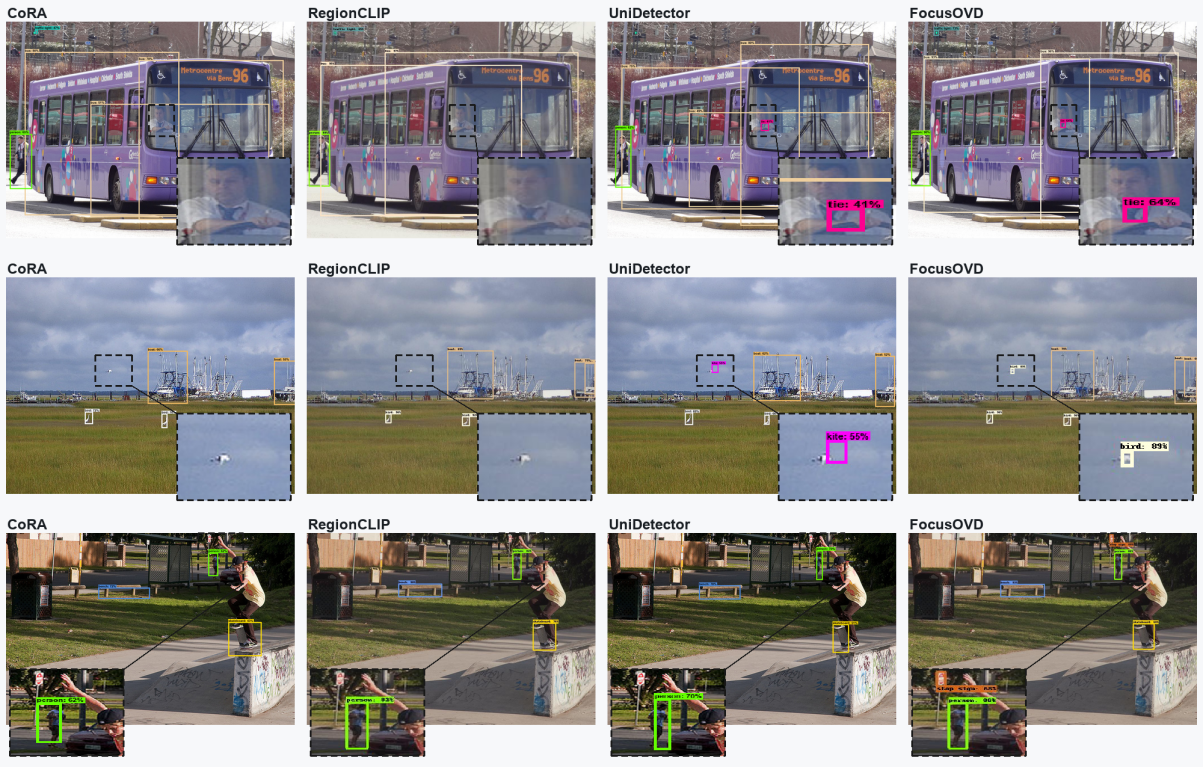}
\caption{Qualitative comparison of CORA, RegionCLIP, UniDetector, and FocusOVD on COCO validation images.}
\label{qualitative comparison}
\end{figure}

To complement the quantitative comparison, Fig.~\ref{qualitative comparison} presents qualitative detection results of CORA, RegionCLIP, UniDetector, and FocusOVD on COCO validation images. The zoomed regions show that FocusOVD can recover several small objects that are missed or poorly localized by the competing methods.

LVIS \cite{ref12} provides a more challenging long-tailed setting. We report overall AP together with AP for rare (\(\mathrm{AP}_{\mathrm{r}}\), fewer than 10 training instances), common (\(\mathrm{AP}_{\mathrm{c}}\), 10--100 instances), and frequent (\(\mathrm{AP}_{\mathrm{f}}\), more than 100 instances) categories. The rare-category score is particularly informative for evaluating open-vocabulary generalization.

\begin{table}[H]
\caption{Open-vocabulary detection results on LVIS-OVD.}
\label{tab2}
\centering
\small
\begin{tabular}{lcccc}
\hline
Model & \(\mathrm{AP}_{\mathrm{r}}\) & \(\mathrm{AP}_{\mathrm{c}}\) & \(\mathrm{AP}_{\mathrm{f}}\) & AP \\
\hline
ViLD \cite{ref4} & 16.1 & 23.9 & 32.5 & 24.9 \\
RegionCLIP \cite{ref5} & 21.3 & 30.8 & 36.0 & 30.7 \\
OV-DETR \cite{ref17} & 17.4 & 25.0 & 32.5 & 26.6 \\
Detic \cite{ref18} & 17.8 & 26.3 & 31.6 & 26.8 \\
CoDet \cite{ref10} & 21.8 & 30.2 & 34.9 & 30.6 \\
MEDet \cite{ref11} & \underline{22.2} & \textbf{32.7} & \underline{36.9} & \underline{33.1} \\
CORA \cite{ref8} & 22.0 & 31.4 & 35.8 & 31.3 \\
UniDetector \cite{ref9} & 20.2 & 28.7 & 31.0 & 29.4 \\
FocusOVD & \textbf{23.9} & \underline{32.6} & \textbf{37.1} & \textbf{33.7} \\
\hline
\end{tabular}
\end{table}

As reported in Table~\ref{tab2}, FocusOVD obtains 33.7\% overall AP and the best \(\mathrm{AP}_{\mathrm{r}}\) and \(\mathrm{AP}_{\mathrm{f}}\) among the compared methods, reaching 23.9\% and 37.1\%, respectively. Its \(\mathrm{AP}_{\mathrm{c}}\) of 32.6\% is within 0.1\% of the best result. The consistent performance across category frequencies demonstrates that FocusOVD improves recognition of rare categories without sacrificing detection accuracy on common and frequent categories.

\subsection{Cross-Dataset Zero-Shot Transfer}

We train FocusOVD on the LVIS base categories and evaluate it on the COCO and Objects365 v2 \cite{ref14} validation sets to assess cross-dataset zero-shot transfer. FocusOVD uses the CLIP ResNet-50 backbone in Table~\ref{tab3}. We report AP, \(\mathrm{AP}_{50}\), and \(\mathrm{AP}_{75}\), where AP averages precision over IoU thresholds from 0.50 to 0.95, while \(\mathrm{AP}_{50}\) and \(\mathrm{AP}_{75}\) measure performance at IoU thresholds of 0.50 and 0.75, respectively.

As shown in Table~\ref{tab3}, FocusOVD achieves the best performance on COCO, with 37.0\% AP, 56.0\% \(\mathrm{AP}_{50}\), and 40.1\% \(\mathrm{AP}_{75}\). Its AP exceeds ViLD, F-VLM, DetPro, BARON, and UniDetector by 0.4\%, 4.5\%, 2.1\%, 0.8\%, and 0.6\%, respectively. On Objects365 v2, FocusOVD obtains 13.8\% AP, and achieves the highest \(\mathrm{AP}_{75}\) of 15.0\%. Although its \(\mathrm{AP}_{50}\) of 21.1\% is slightly below the best result, the overall results demonstrate strong transfer across datasets with different category distributions and visual characteristics.

\begin{table}[H]
\caption{Cross-dataset zero-shot transfer results on COCO and Objects365 v2.}
\label{tab3}
\centering
\small
\begin{tabular}{lcccccc}
\hline
\multirow{2}{*}{Model} & \multicolumn{3}{c}{COCO} & \multicolumn{3}{c}{Objects365 v2} \\
\cline{2-7}
 & AP & \(\mathrm{AP}_{50}\) & \(\mathrm{AP}_{75}\) & AP & \(\mathrm{AP}_{50}\) & \(\mathrm{AP}_{75}\) \\
\hline
ViLD \cite{ref4} & \underline{36.6} & 55.6 & \underline{39.8} & 11.8 & 18.2 & 12.6 \\
F-VLM \cite{ref7} & 32.5 & 53.1 & 34.6 & 11.9 & 19.2 & 12.6 \\
DetPro \cite{ref19} & 34.9 & 53.8 & 34.7 & 12.1 & 18.8 & 12.9 \\
BARON \cite{ref20} & 36.2 & \underline{55.7} & 38.1 & \underline{13.6} & 21.0 & 14.5 \\
UniDetector \cite{ref9} & 36.4 & 54.9 & 38.5 & 13.3 & \textbf{21.4} & \underline{14.7} \\
FocusOVD & \textbf{37.0} & \textbf{56.0} & \textbf{40.1} & \textbf{13.8} & \underline{21.1} & \textbf{15.0} \\
\hline
\end{tabular}
\end{table}

\subsection{Ablation Study}

We train FocusOVD on the LVIS base categories and evaluate it on the COCO validation set to isolate the individual and combined effects of RCFF, RPGConv, and RCGLU. All ablation configurations use the CLIP ResNet-50$\times$4 backbone. Table~\ref{tab4} reports the results of the component ablation study; values in parentheses denote absolute gains over the baseline without the proposed modules.

\begin{table}[H]
\caption{Component ablation results.}
\label{tab4}
\centering
\small
\setlength{\tabcolsep}{4pt} 
\resizebox{\textwidth}{!}{
\begin{tabular}{cccccccccc}
\hline
\# & RCFF & RPG & RCGLU & AP & \(\mathrm{AP}_{50}\) & \(\mathrm{AP}_{75}\) & \(\mathrm{AP}_{S}\) & \(\mathrm{AP}_{M}\) & \(\mathrm{AP}_{L}\) \\
\hline
1 & -- & -- & -- & 35.5 & 56.9 & 38.2 & 21.6 & 39.0 & 52.8 \\
2 & \checkmark & -- & -- & \begin{tabular}[c]{@{}c@{}}38.8 \\ \footnotesize (+3.3)\end{tabular} & \begin{tabular}[c]{@{}c@{}}60.4 \\ \footnotesize (+3.5)\end{tabular} & \begin{tabular}[c]{@{}c@{}}41.2 \\ \footnotesize (+3.0)\end{tabular} & \begin{tabular}[c]{@{}c@{}}23.2 \\ \footnotesize (+1.6)\end{tabular} & \begin{tabular}[c]{@{}c@{}}42.3 \\ \footnotesize (+3.3)\end{tabular} & \begin{tabular}[c]{@{}c@{}}56.9 \\ \footnotesize (+4.1)\end{tabular} \\
3 & \checkmark & \checkmark & -- & \begin{tabular}[c]{@{}c@{}}\underline{39.5} \\ \footnotesize (+4.0)\end{tabular} & \begin{tabular}[c]{@{}c@{}}60.8 \\ \footnotesize (+3.9)\end{tabular} & \begin{tabular}[c]{@{}c@{}}\underline{42.8} \\ \footnotesize (+4.6)\end{tabular} & \begin{tabular}[c]{@{}c@{}}24.7 \\ \footnotesize (+3.1)\end{tabular} & \begin{tabular}[c]{@{}c@{}}43.4 \\ \footnotesize (+4.4)\end{tabular} & \begin{tabular}[c]{@{}c@{}}\underline{57.5} \\ \footnotesize (+4.7)\end{tabular} \\
4 & \checkmark & -- & \checkmark & \begin{tabular}[c]{@{}c@{}}39.2 \\ \footnotesize (+3.7)\end{tabular} & \begin{tabular}[c]{@{}c@{}}\underline{61.6} \\ \footnotesize (+4.7)\end{tabular} & \begin{tabular}[c]{@{}c@{}}42.4 \\ \footnotesize (+4.2)\end{tabular} & \begin{tabular}[c]{@{}c@{}}\underline{25.1} \\ \footnotesize (+3.5)\end{tabular} & \begin{tabular}[c]{@{}c@{}}\underline{43.7} \\ \footnotesize (+4.7)\end{tabular} & \begin{tabular}[c]{@{}c@{}}57.2 \\ \footnotesize (+4.4)\end{tabular} \\
5 & \checkmark & \checkmark & \checkmark & \begin{tabular}[c]{@{}c@{}}\textbf{39.9} \\ \footnotesize (+4.4)\end{tabular} & \begin{tabular}[c]{@{}c@{}}\textbf{62.1} \\ \footnotesize (+5.2)\end{tabular} & \begin{tabular}[c]{@{}c@{}}\textbf{44.6} \\ \footnotesize (+6.4)\end{tabular} & \begin{tabular}[c]{@{}c@{}}\textbf{26.4} \\ \footnotesize (+4.8)\end{tabular} & \begin{tabular}[c]{@{}c@{}}\textbf{44.5} \\ \footnotesize (+5.5)\end{tabular} & \begin{tabular}[c]{@{}c@{}}\textbf{57.9} \\ \footnotesize (+5.1)\end{tabular} \\
\hline
\end{tabular}
}
\end{table}

The baseline achieves 35.5\% AP and 21.6\% \(\mathrm{AP}_{S}\). Adding RCFF increases these scores to 38.8\% and 23.2\%, corresponding to gains of 3.3\% and 1.6\%. Adding RPGConv to RCFF yields 39.5\% AP and 24.7\% \(\mathrm{AP}_{S}\), corresponding to gains of 4.0\% and 3.1\% over the baseline. Replacing RPGConv with RCGLU in the RCFF-based configuration yields 39.2\% AP and 25.1\% \(\mathrm{AP}_{S}\), corresponding to gains of 3.7\% and 3.5\% over the baseline, respectively. With all three modules enabled, FocusOVD achieves 39.9\% AP and 26.4\% \(\mathrm{AP}_{S}\), improving over the baseline by 4.4\% and 4.8\%. It also gains 5.2\%, 6.4\%, 5.5\%, and 5.1\% in \(\mathrm{AP}_{50}\), \(\mathrm{AP}_{75}\), \(\mathrm{AP}_{M}\), and \(\mathrm{AP}_{L}\), respectively. The progressive improvements indicate that the three modules provide complementary benefits across cross-scale fusion, local geometric modeling, and RoI feature refinement.

\section{Conclusion}

In this paper we introduced FocusOVD, an open-vocabulary detector designed for small objects. It enhances frozen vision-language features through RPGConv for local geometric modeling, RCFF for cross-scale fusion, and RCGLU for region-level refinement. Experiments on COCO, LVIS, and Objects365 demonstrate competitive performance in open-vocabulary detection and cross-dataset transfer, particularly for small objects. Future work will extend this feature-refinement framework to open-vocabulary instance segmentation and other dense prediction tasks.
%
%
%
%

\end{document}